# Modeling Intelligent Decision Making Command And Control Agents: An Application to Air Defense

Sumanta Kumar Das

Institute for Systems Studies and Analyses, Delhi, India

*Abstract*— The paper is a half-way between the agent technology and the mathematical reasoning to model tactical decision making tasks. These models are applied to air defense (*AD*) domain for command and control (*C2*). It also addresses the issues related to evaluation of agents. The agents are designed and implemented using the agent-programming paradigm. The agents are deployed in an air combat simulated environment for performing the tasks of *C2* like electronic counter counter measures, threat assessment, and weapon allocation. The simulated *AD* system runs without any human intervention, and represents state-of-the-art model for *C2* autonomy. The use of agents as autonomous decision making entities is particularly useful in view of futuristic network centric warfare.

## I. INTRODUCTION

In recent times information sharing and collaborative decision making over the defense networks have completely revolutionized the air combat operations [1]. Today's offensive forces are equipped with sophisticated electronic attacking (*EA*) or electronic counter measuring (*ECM*) devices (designed to interfere radar and communication systems), airborne warning and controlling system (*AWACS*) aircrafts, high precision air-to-air, air-to-surface missiles, high speed fighters, bombers, unmanned air vehicles (*UAV*) etc. To respond to these, the defensive forces rely on early warning surveillance or tracking radar that has electronic counter counter measures (*ECCM*), high-tech command and controls (*C2*) that robustly assess the threats and efficiently allocate right weapons for engaging right targets. Modeling such decision making *C2* is of utmost importance to survive with such technological advancement.

An integrated air defense (*AD*) system is an aggregation of sensors, weapons, *C2*, intelligence systems, communications, and personnel operating under the command of a designated *AD* commander. The *AD* systems have progressed steadily over the recent years to include highly sophisticated computer-based software systems to assist and train the commander. Some of the examples of such tools are Air Force Mission Support System, PowerScene, TopScene etc. [1].

Usually decision making processes of *C2* involve an *OODA* (Observe-Orient-Decide-Act) loop or variants of it ([2], [3]). Although, the *OODA* loop was initially originated from behavioral science, latter it was exploited for understanding the human participations in complex *C2* problems. Along with the *OODA* loop recently, the *BDI* (Belief- Desire-Intention) architectures of agent oriented approach are also becoming popular because of it′s enhanced capability of practical reasoning for developing intelligent software systems. It has the advantages from the user perspective in terms of both speed and ease of development of models.

Any *AD* system is highly dependent on classifying targets, doing intent recognition and threat assessment (*TA*). Several multidisciplinary studies have been performed to solve such problems. The multiple attribute decision making has been applied for *TA* in [4]. In [2, 5, and 6], threat is assessed in terms of risk using dynamic Bayesian networks which requires the knowledge of prior probabilities. Since this information is not always available, alternative approaches such as fuzzy logic is seldom found to be useful.

Today, most of the sensors and weapons of *AD* systems can perform multiple tasks or engagements. Operating such systems autonomously is a challenging task. In past, decision making process for such *C2* are modeled and automated using fuzzy logic, dynamic Bayesian networks, decision trees [7], neural networks [8] etc. In those studies, *TA* and decision making processes are modeled separately. In this study, an integrated approach of *TA* prior to decision making (weapon allocation (*WA*)) based on the *BDI* architectures of agent modeling is proposed.

Looking at the real applications of agent technologies starting from the sensor networking [9] to coordinating ambulances in emergency situation [10] or adaptive traffic control [11], one can think of applying these technologies to *C2* processes of *AD* system. Technologies like normality analysis [12] to automatically understand complex environment and to detect abnormal behavior of events of interest can also be brought into the *C2* system for situational assessment.

This paper is mainly focused on two aspects, firstly on the modeling the *C2* of *AD* system in terms of *BDI* architectures, and secondly, evaluating the system on the basis of correct decisions in a simulated environment and by the opinion of human operators. Two decision making software agents are discussed. First one is related with electronic counter counter measures (*ECCM*) against electronic jamming and second one is related with *TA* and *WA*.

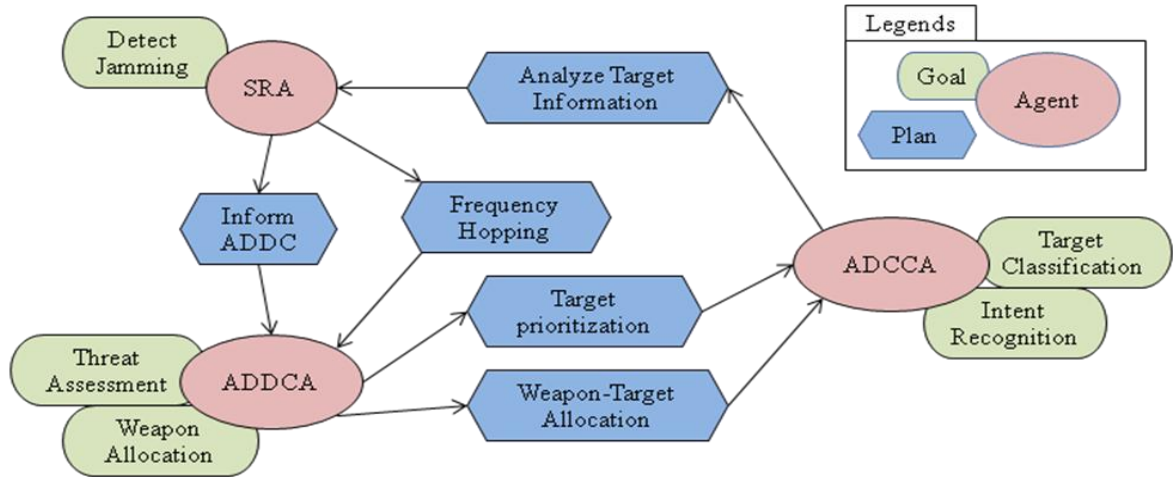

Figure 1. Requirements analysis of multi-agent air defense system using the TROPOS concepts.

## II. APPLICATION IN AIR DEFENSE

Let us assume an autonomous *AD* squadron, composed of several smaller units (batteries) with variable *AD* capabilities (like short- medium- or long-range, low-to-high-altitude, air and ground missile defense systems, *AD* guns etc., (e.g. an *AD* squadron composed of *Patriot* (long range missile), Hawk *XXI* (medium range missile), or *NASAMS* (*s*hort or medium range missile)), is deployed to defend a vulnerable area or point (*VAVP*) (e.g. runways, tank platoons etc.), against a point, area or maneuver air attack. Batteries are equipped with their own surveillance radar, tracking radar and command post (Air Defense Direction Center or *ADDC*). Central *C2* is governed by squadron level headquarter (Air Defense Command Center or *ADCC*). For early warning, *ADCC* relies on central acquisition radar (*CAR*). The *CAR* starts tracking targets at larger distance (around 200 km). Track information is transferred to *ADCC*. The *ADCC* classify targets and passed information to *ADDC*. Surveillance radar starts tracking target at lesser range (around 100 km). This data is transferred to *ADCC*. The *ADCC* performs multi radar tracking and carries out track correlation and data fusion and send the target-list to *ADDC*. The *ADDC* prioritizes threats from selected list of targets (which may include aircrafts, helicopters, tactical ballistic missiles, cruise missiles, *UAV*s etc.). The *ADDC* then perform optimal assignment of available weapons to targets. A multi-agent system (*MAS*) can be designed comprising of *ADDC*, *ADCC* and surveillance radar as agents. These agents depend on each other for achieving their goals. Figure 1 shows the requirements analysis of *MAS* using TROPOS (a tool used for designing agent-based systems) [13].

### A. Electronic Counter Counter Measures(ECCM):

Today′s aircrafts (e.g. *Su-25*) are equipped with *EA* systems like *jamming* to reduce effectiveness of radars. Significant changes in surveillance sectors give an indication of jamming. The radar operator is capable of identifying whether at given time the enemy is using its jammer or not from received information like "target status ($n_t$)" that includes target′s position, altitude, speed etc at time *t*.

On the basis of $n_t$ radar operators decides their actions. The relative difference (*RD*) of target status ($|((n_t-n_{t+1})/n_t)|$) gives an indication of jamming. If the degree of jamming is very high usually the radar operator sends message to *ADDC*. On getting this information *ADDC* allocates interceptor aircraft to investigate about target. The *ADDC* sends this massage to the pilot for investigating suspected targets. The pilot prioritizes and engages targets by air-to-air missile based on its capability and availability and sensor performance (detection range). A heterogeneous range of sensors from ground as well as airborne are assigned to suspected target for tracking.

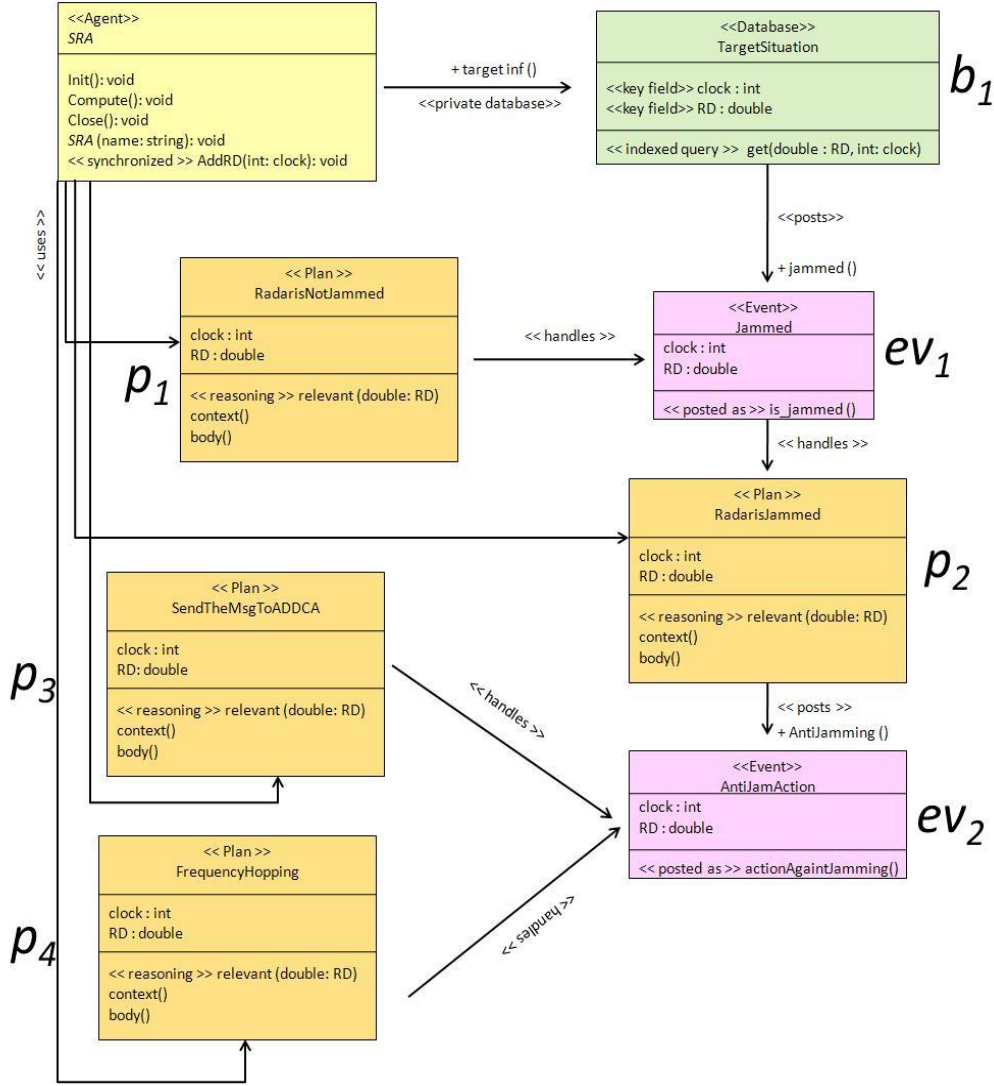

Figure 2. Class diagram of a surveillance radar agent that performs the task of *ECCM* and has been implemented in the *JACK*-agent oriented programming language. Each box represents a class stereotype, it has three parts containing the class label, attributes and methods. The arrows between the classes represent the association. The notations *b*, *ev* and *p* are used to represent the agent's beliefset, event and plan respectively.

All detected targets are not necessarily hostile. Targets are electronically identified using predefined codes. Predefined codes are distributed among friendly units. If the incoming unit responds correctly to these codes, it is regarded as friendly. If the response is contrary, the unit is considered to have "suspect" and thus invites tracking.

B. *Threat Assessment And Weapon Allocation*

i. *Threat Assessment (TA) :*

For *TA* fuzzy inference rules are used [14]. These rules are required to be stored in the agent's beliefsets. The fuzzy inference rules have following form:

**if** $A_1$ is $S_1$ and … and $A_n$ is $S_n$ **then** $F$ is $L_1$

$A_i$ and $F$ are fuzzy variables and $S_j$ and $L_1$ are fuzzy labels. $A_i$'s are the input variables; $F$ is the output variable. For each fuzzy variable, fuzzy labels are defined as follows:

*Inputs*
- *Range :* { Close, Medium, Far },
- *Velocity:* {Slow, Medium, Fast },
- *Altitude:* {Low, Medium, High} ,
- *Angle of Attack:* {Low, Medium, High },
- *Targets Types:*{Very Lethal, Lethal, Less Lethal},

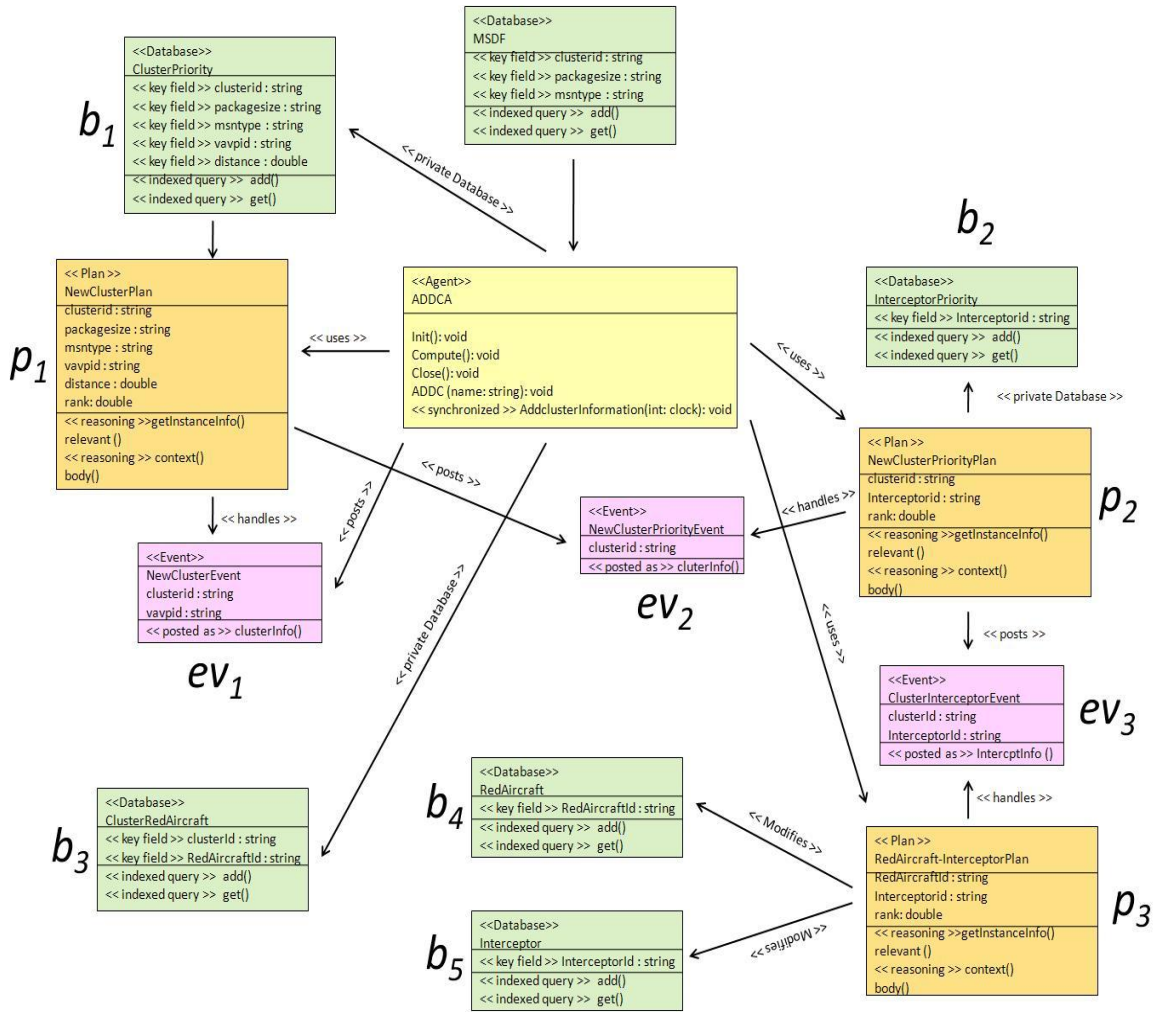

Figure 3. Class diagram of an *ADDC* agent that performs the tasks of threat prioritization and weapon allocation and has been implemented in *JACK*-agent oriented programming language.

- *Intent Classes:* {Strike, Interdiction, Suppression, Tactical Bombing, Strategic Bombing, Electronic, Close Air Support, Escort, Surveillance, Reconnaissance}.

*Output*
- *Threat:* {Low, Medium, High}.

The angle between the target′s velocity vector projection and the longitudinal axis is defined as the angle of attack (*AOA*). The intent classes membership values are function of an operational parameter called conflict level (*CL*) [2]. This *CL* represents seriousness of a situation, $0 \leq CL \leq 1$, *CL*= 0 indicates peace time and *CL*=1 indicates full scale war. The value of *CL* is given by users based on their assessment of the situation to the system. The Target Type is divided into three fuzzy sets, very lethal (missile, a group of bomber or fighter), lethal (a fighter or a bomber) and less lethal (*EA*, *AWACS* and other aircrafts).

The fuzzy system consists of fuzzy rules such as :

- *R1*: If the target′s *Range* is Far and *Velocity* is Slow and *Altitude* is High and *Angle of Attack* is Low and *Target Type* is Less Lethal and *Intent Class* is Reconnaissance or Surveillance, then its *Threat* is Low.

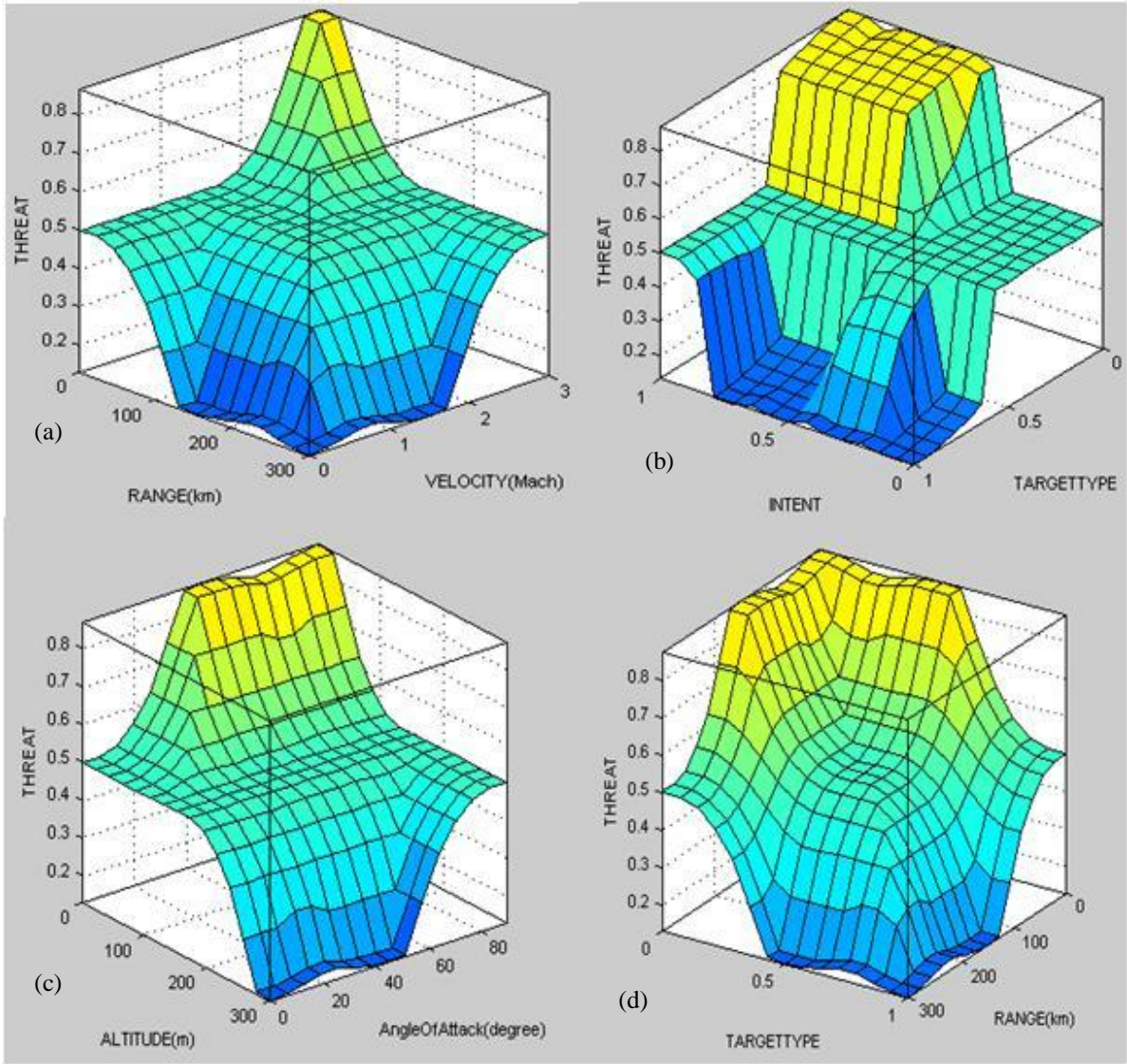

Figure 4. Surface plots of fuzzy inference rules for prioritizing threats. The figures show the variations of threat as a function of (a) *range* and *velocity* (b) *intent* and *target types* (c) *altitude* and a*ngle of attack* and (d) *target type* and *range* while keeping other factors as fixed variable.

- *R1215*: If target′s *Range* is Close and *Velocity* is Fast and *Altitude* is Low and *Angle of Attack* is High and *Target Type* is Very Lethal and *Intent Class* is Strike or Interdiction, then its *Threat* is High.

These rules are written on the basis of intuitive and expert considerations and then tuned by simulation tests. A Mamdani approach is followed. The input/output fuzzy sets are defined using trapezoidal and semi-trapezoidal membership functions. The 'and' operator and the implication methods are the product, and the defuzzification method is weighted average. Total 1215 possible fuzzy inference rules ($=3^5 \times 5$) are possible, but all rules need not to be defined because few of the rules unlikely to be observed in real situation. A minimal set of rules (e.g. a heuristic is defined as in rules of *R1* and *R1215* that represent two extreme situations of threat perception) are defined and other possible rules are automatically interpolated through given minimal set of rules. The ranking of each rules are performed in agent's plan-base.

*ii. Weapon Allocation (WA):*

Since 1959, *WA* problem has been extensively studied in operations research for further improvement [15]. Recently evolutionary approaches are found effective for *WA* [16]. Weapons are allocated to attacking targets based on target types, weapons' effectiveness, range, and availability. An integer linear programming model is developed. The objective function for *WA* is to maximize the target-value-destroyed (*TVD*), which is defined as:

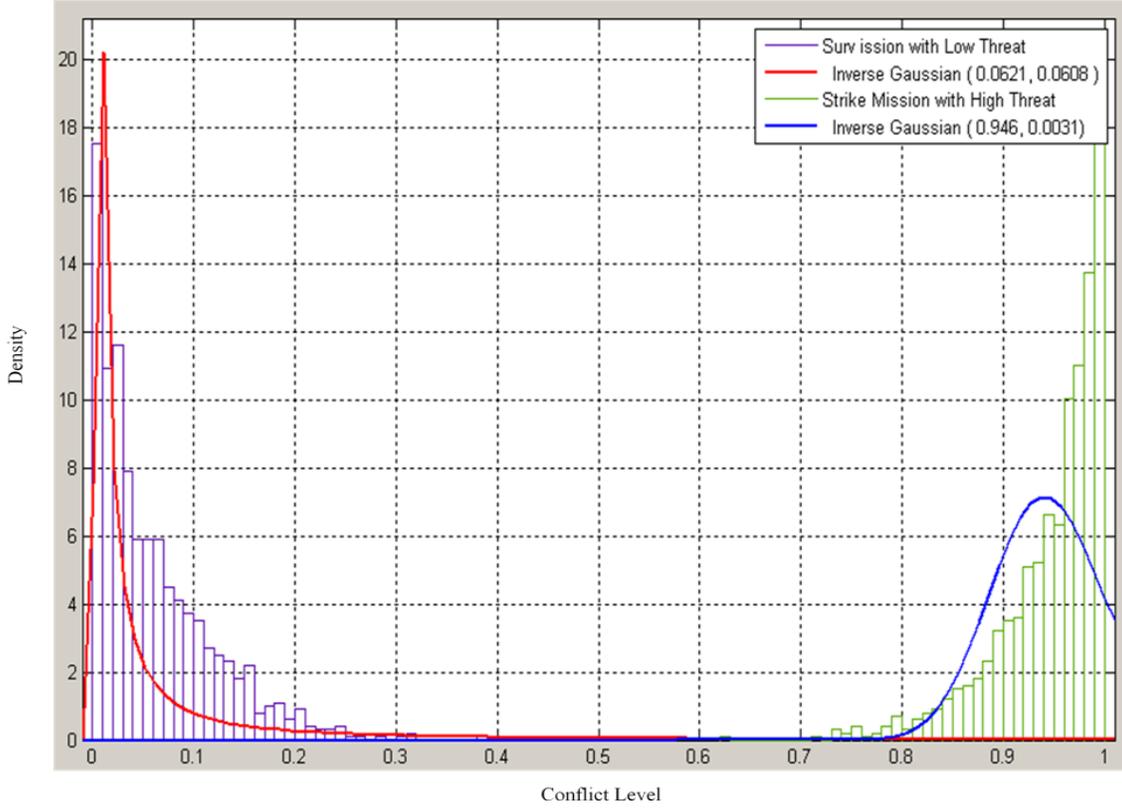

Figure 5. Response of an *ADDCA* based on the *KS* statistics. In an *AD* scenario target may be in any situation ranging from low (far range, high altitude, slow velocity, low *AOA*, less lethal and surveillance or reconnaissance intent type) to high (close range, low altitude, fast velocity, high *AOA*, very lethal type and strike or interdiction intent type) threat label. Assuming a *uniform (0, 1)* distribution of occurrences of these situations the distribution of minimum (low threat) and maximum (high threat) order statistics are found using the *KS* test. The *Inverse Gaussian* distribution is found to be the best fit for both the cases with parameters given in the parenthesis.

$$TVD = \sum_{w=1}^{W}\sum_{s=1}^{S} y_{ws} C_{ws} \quad (1)$$

where $y_{ws}$ represents number of weapons of type $w$ ($w=1,..,W$) allocated to target type $s$ ($s=1,..,S$). $C_{ws}$ is constant matrix whose elements are determined as a function of kill probability and the constraints are (2) $\sum_{w=1}^{W} y_{ws} N_{ws} \leq NI_w$, (3) $\sum_{w=1}^{W} y_{ws} \leq NT_s$ and (4) $y_{ws} \geq 0$, where $N_{ws}$ is number of $w^{th}$ type of weapon required to kill $s$ type target, $NI_w$ is number of weapon of type $w$ and $NT_s$ is number of targets of type $s$.

The models of *C2*-agents are implemented in the *JACK* [17] agent oriented programming language. Agents in *JACK* are autonomous software components that have explicit goals (desires) to achieve or events to handle. The intentions of those agents are modeled as goals. The goals are achieved through execution of some plans and furnishing some resources. Figures 2 and 3 show details of agent's structures using class diagrams (as one representative of unified modeling language). Figure 2 shows that *SRA* has to execute plans $p_1$ to $p_4$ to achieve the intention of *ECCM*. It uses the resources $b_1$ to $b_5$ which generates the events ($ev_1$, $ev_2$) that are prerequisite to execute plans. Similarly, figure 3 shows that the intention of *TA* and *WA* of *ADDC*-agent (*ADDCA*) is achieved by executing plans $p_1$ to $p_3$ and using resources $b_1$ to $b_5$ which generates events ($ev_1$ to $ev_3$) as prerequisites to execute plans. The agents use meta-level plan reasoning (*MPR*) process, as described in [18], for optimal decision making. The *JACK* software provides various library functions (e.g. *relevant*, *context*, *getInstanceInfo* etc.) for *MPR*. The *relevant* function is used to select plan associated with particular event. The *context* function is used to select the plan which is consistent with the agent's current beliefs. If there are still multiple plans left in the applicable plan-base after using *relevant* and *context* functions, the *JACK* provides the *getInstanceInfo* function, which returns a *PlanInstanceInfo* object. This class has the *def* method which returns the rank (or threat value) of the plan. The ranking of each plan is computed by fuzzy rules. Each fuzzy rule generates a distinct plan. The plan with maximum rank is selected by the *getInstanceInfo* function. The events are posted either by the agent itself or by other plans. The events are posted by the plan (e.g. *NewClusterPriorityEvent*, in Figure 3) when the associated plans are executed. This way, most potential weapon is allocated to the most threaten target. While allocating a weapon to the target the agent also checks its availability status so that multiple allocations do not take place. If there is slight change in the calculation process of threat ranking, *MPR* will take care of that. The *MPR* will always select the plan that measures maximum rank.

## III. EVALUATION

This paper proposes two modeling approaches for evaluating the agent's models, namely, logical and statistical. The propositional logic is used for the logical and techniques of hypothesis testing are used for the statistical evaluation.

### A. Logical Evaluation :

The rules based on the propositional logic are used for representing the agent's goals and their mutual conflicts. If an agent pursues multiple goals then the rules should automatically detect it as conflict situation. For the *SRA*, let the mutually exclusive goals (or plans) are $p_1$: *RadarIsNotJammed*, $p_2$: *RadarIsJammed*, $p_3$: *SendTheMessageToADDCA*, $p_4$: *Frequency Hopping*. The rules are defined as:

$$\{ev_1 \in Jammed\}^{\beta}, \{p_1\}^{k-} \Rightarrow \{p_2 \vee p_3 \vee p_4\} \quad (5)$$
$$\{ev_1 \in \neg Jammed\}^{\beta}, \{p_2 \vee p_3 \vee p_4\}^{k-} \Rightarrow p_1 \quad (6)$$
$$\{ev_2 \in \neg Jammed\}^{\beta}\{p_2\}^{k+} \Rightarrow \{p_3 \vee p_4\} \quad (7)$$

where *ev's* represents events, posted by the agent when it percepts significant changes in the environment. The equation *(5)* represents that if the *SRA* is *Jammed* (belief state denoted by $\beta$), it may derive the goal to go for $p_2$ or (denoted by $\vee$) $p_3$ or $p_4$, but the goal to go for $p_1$ is in conflict (denoted by *k-*). Similarly, the equations *(6)* and *(7)* also define the agent's operating conditions and their conflicts. Other possible conflicting goals situations are $\{p_3 \wedge p_4\}$, $\{p_1 \wedge p_2\}$, $\{p_3 \wedge p_1\}$, $\{p_4 \wedge p_2\}$, because if a *SRA* is found jammed, it may go for the plan $p_3$ or $p_4$, but should not pursue these goals (or plans) simultaneously though these are mutually exclusive plans.

The state transition of agent is obtained at each time step for each plan instances ($p_1$ to $p_4$). The output of each time instance is a vector containing four strings representing the agent's state $< p_1, p_2, p_3, p_4 >$. After five hundreds simulation runs no conflicting goal situation is observed for the *SRA*.

Similarly, for the *ADDCA*, the plans and the conflicting situations are defined as follows:

$p_{1(N1)}$: *NewClusterPlans*,
$p_{2(N2)}$: *NewClusterPriorityPlans*,
$p_{3(N3)}$: *RedAircraft-InterceptorPlans*,

where each plan $p_i$'s may posses a plan-base of *Ni* plans.

$$\{ev_1 \in max(rank(p_{1(N1)}))=p_{1(i)}\}^{\beta}, \{p_{1(j), j \neq i}\}^{k-} \Rightarrow \{p_{1(i)}\} \quad (8)$$
$$\{ev_2 \in max(rank(p_{2(N2)}))=p_{2(i)}\}^{\beta}, \{p_{2(j), j \neq i}\}^{k-} \Rightarrow \{p_{2(i)}\} \quad (9)$$
$$\{ev_3 \in max(rank(p_{3(N3)}))=p_{3(i)}\}^{\beta}, \{p_{3(j), j \neq i}\}^{k-} \Rightarrow \{p_{3(i)}\} \quad (10)$$

If the $i^{th}$ plan combination has maximum rank then the above rules are true.

### B. Statistical Evaluation:

The null hypothesis ($H_0$) is assumed that the agent's behavior (a random variable $X$) be characterized by a fully specified statistical distribution $F(x)$. The Kolmogrov-Smirnov (*KS*) statistic is used to determine the goodness-of-fit of the underlying distribution pattern of the agent's behavior. The *KS* statistic is defined as $D_n = sup_x(|F_n(x)-F(x)|)$, where $F_n(x)$ is the empirical distribution function of random variable $X$ of a sample of size $n$. The hypothesis about the distributional form is rejected at the chosen significance level ($\alpha$) if the test statistic, *KS*, is greater than the critical value obtained from statistical table. The *KS* statistic is applied for a number of statistical distributions and a ranking is performed for all of the fitted distributions. The fitted distribution with the highest *KS* rank is being selected as the characterized distribution and the performance measure of the agent.

The system is designed in such a way that we need not to specify the statistical distribution in $H_0$. The system automatically decides the best fitted distribution with estimated parameters from a library of distributions on the basis of *KS* statistic.

## IV. RESULTS AND DISCUSSION

An air combat scenario of smaller scale (200 km × 200 km) is simulated where offensive force has one ground-attack aviation regiment composed of one squadron (10 aircrafts) of high speed fighter (e.g. *A-10 Thunderbolts*) and bomber (e.g. *F-117*) each, 10 air-to-surface missiles (*Maverick*), 15 cruise missiles (e.g. *Tomahawk*), 50 smart bombs, one *UAV* and one *AWACS* aircraft. The force is using electronic jammer (like directed energy into the enemy's search radar) for *EA*. Each unit of this force is approaching from different directions (with different speeds, altitudes and ranges), simultaneously towards a *VAVP* (a runway and aircraft shelters), which is protected by one squadron of integrated *AD* system comprising of one surveillance radar (capable of *ECCM*), one tracking radar, interceptor aircrafts two batteries (each with 3 units) of long (e.g. *Patriot*), medium (e.g. Hawk *XXI*) and small (e.g. *NASAMS*) range *SAMs* and Anti-Aircraft Artillery and one agent based *C2* system.

Because of jamming, the surveillance radar receives wrong measurements of $n_t$ at time $t$. The $n_t$ is generated randomly using the Poisson distribution with mean 20. The *RD* values are computed at each time step and found that it best fits to *General extreme value* distribution with respect to the *KS* statistics. The calculated value of the *KS* statistic (0.6623) is greater than the theoretical value (0.136685 at confidence level *95%*) so the $H_0$ that the *RD* follows the specified distribution (*General extreme value*) is accepted and gives an indication of jamming and performance measure of the agent. It is observed that out of 500 runs, 231 times (i.e. 41 %) the radar is found to be jammed and 91 times the radar is found to send the message to the *ADDCA*.

The *ADDCA* starts prioritizing once the targets reach within 200 km range from *VAVP*. Principal findings of the simulation results suggest that if fast moving very lethal target type (a group of fighter *A-10 Thunderbolts* with speed 2.5 Mach) is very close (within 100 km) to the *VAVP*, its priority is very high as compared to a relatively slow moving target (*Tomahawk* missile with speed 0.7 Mach) which is quite far (beyond 200 km) (Fig 4(a)). Also if a lethal target (a group of bomber *F-117*) is coming with strike intention then its priority is more than a relatively less lethal target (*EA* aircraft) is coming with reconnaissance intention (Fig 4(b)). Also a target in a very low altitude (*Su-27* in a *SEAD* (Suppression of Enemy *AD*) mission) and high angle of attack is very dangerous than a target in high altitude moving in low angle of attack (*UAV*) (Fig 4(c)). Similarly, the threat of a low lethal target type (cargo aircraft) with the intention of attacking the *VAVP* (asymmetric warfare) at a very close distance is very high than a very lethal target at far range (*Su-27*) (Fig 4(d)).

The basic goal of the *ADDCA* is to prioritize threats based on estimated target′s status. For example, let a plan detects target with far range, high altitude, slow velocity, low *AOA*, less lethal type and surveillance intention then it should assign low threat label, similarly, if a plan detects target with close range, low altitude, fast velocity, high *AOA*, very lethal type and strike intention then it should assign the high threat label. So the former plan instance is considered to follow the distribution of minimum order statistics and later one be considered to follow the distribution of higher order statistics. Now the distribution of these order statistics depends on the distribution of their parents. Assuming a uniform parent distribution to all plan instances the resulting distribution of minimum and maximum order statistics are shown in figure 5. The *KS* statistics are measured from the distributions of order statistics and the *Inverse Gaussian* distribution is found to be the best fit for both the cases.

The *SRA* requires lesser computation time for checking its logical conditions than the *ADDCA*. The *SRA* computes the *RD* based on the current and previous observations; therefore, it requires at least two observations. Once it calculates the *RD* it checks the threshold $\alpha$. Similarly, for the *ADDCA,* planning time is fixed to 30 seconds. A genetic algorithm is developed for searching the optimal conditions. As expected, the search quality decreases with increasing number of targets. The decrease is not just because of the increase of the planning problem complexity, but also, and most importantly, because the number of available defensive combat resources and their configuration are kept fixed, for more targets to defend against. The system is tested for 30 weapons against 90 targets, up to this level the system performs effectively, above this level its performance reduces. So for larger operations more than one *C2* systems have to be integrated.

The reasoning process of the system is explained to the users and they found it logical and its conclusions sound to them relevant and useful. In future the plan is to conduct exhaustive experimentations of the system with diverse *AD* scenarios to extend the model for identifying the friend and foe as a function of *CL* and other rules of engagement (like visual identity) to further increase the user acceptance and usability. Also the proposed system will be evaluated with realistic scenarios (e.g. *Operation Desert Storm).* The aerial operations similar to those scenarios will be generated and the beliefs and decisions of the agents will be evaluated. This will help to develop air war-games where experts can evaluate the systems by changing their plans and strategies. For simplicity few operational choices used in this study are straight forward (e.g. target type, intent classes), in future these will be addressed more thoroughly. This system is conceptualized in computer simulated environment; the issues related to operational-level analysis can only be addressed in future.

Similarly, inclusion of soft kill or non-lethal, options like decoys, chaffs, relocation of *AD* forces, deterrence measures, jamming etc. are left for future considerations. Further tests are to be done in future using two or more *ADDCAs* to see how they may negotiate for optimal utilization of their resources. To protect the system against byzantine attack the agents will communicate with each other with signed (or coded) messages which will be difficult to be forged by the traitor agents. Construction of the signature function will be a cryptography problem and addressed in future.

V. CONCLUSIONS

In this paper, modeling of autonomous intelligent agents for an *AD* system is presented using the concept of meta-level plan reasoning of *BDI* (Belief-Desire-Intention) architectures. The *C2*-agents take decisions of *ECCM*, *TA* and *WA*. The *SRA* decides when to change its frequency to defend the radar system against jamming. *ADDCA* performs *TA* and *WA*. The agents' logic is first formulated in the form of *BDI* architectures and then implemented using the *JACK* agent programming language. The behavioral patterns of the agents in different simulated environments are also presented.

## Acknowledgments

Thanks to the anonymous reviewers for their insightful comments, this significantly improved this article's quality.

Sumanta Kumar Das is working as scientist at Institute for Systems Studies And Analyses, Delhi, India. Contact him at sumantadas.delhi@gmail.com.